\documentclass[ctexart, utf8, 11pt]{article}
\usepackage{nodalida2021}
\usepackage{times}
\usepackage{soul}
\usepackage{epstopdf}
\usepackage{booktabs}

\usepackage{comment}
\usepackage{graphicx}
\usepackage{CJKutf8}  
\usepackage[utf8]{inputenc}

\usepackage{url}
\usepackage{hyperref} 

\usepackage{latexsym}

\usepackage{pgfplots}
\pgfplotsset{compat=1.14}
\usepackage{pgfplotstable}

\definecolor{aqua}{rgb}{0.0, 1.0, 1.0}

\newlength\graphheight
\newlength\graphwidth
\pgfplotsset {
graph/.style={
height=\graphheight,
width=\graphwidth,
legend style={
    draw=none,
    fill=none,
    font=\footnotesize,
},
ticklabel style={font=\scriptsize, /pgf/number format/fixed},
scaled y ticks = false,
scaled x ticks = false,
xlabel near ticks,
ylabel near ticks,
},
graphright/.style={
graph,
yticklabel pos=right,
},
}

\aclfinalcopy 

\title{Chinese Character Decomposition for Neural MT with Multi-Word Expressions}

\author{
           Lifeng Han$^1$, Gareth J. F. Jones$^1$,  Alan F. Smeaton$^2$ \and Paolo Bolzoni\\

         $^1$ ADAPT Research Centre  \\ 
         $^2$ Insight Centre for Data Analytics 
         \\ School of Computing, Dublin City University, Dublin, Ireland \\
         {\tt lifeng.han@adaptcentre.ie}, {\tt paolo.bolzoni.brown@gmail.com} \\}

\date{}

\begin{document}
\maketitle
\begin{abstract}
Chinese character decomposition has been used as a feature to enhance Machine Translation (MT) models, combining radicals into character and word level models. Recent work has investigated ideograph or stroke level embedding. However, questions remain about different decomposition levels of Chinese character representations, radical and strokes, best suited for MT. To investigate the impact of Chinese decomposition embedding in detail, i.e., radical, stroke, and intermediate levels, and how well these decompositions represent the meaning of the original character sequences, we carry out analysis with both automated and human evaluation of MT. Furthermore, we investigate if the combination of decomposed Multiword Expressions (MWEs) can enhance the model learning. MWE integration into MT has  seen more than a decade of exploration. However, decomposed MWEs has not previously been explored.
  
\end{abstract}

\section{Introduction}
Despite Neural Machine Translation (NMT) \cite{cho2014encoder-decoder,Google2016MultilingualNMT,google2017attention,bert4mt2019lample} having recently replaced Statistical Machine Translation (SMT) \cite{brown-etal-1993-mathematics,Och2003CL,chiang-2005-hierarchical,Koehn2010} as the state-of-the-art, research questions still remain, such as how to deal with \textit{out-of-vocabulary} (OOV) words, how best to integrate \textit{linguistic knowledge} and how best to correctly translate \textit{multi-word expressions} (MWEs) \cite{Sag2002MWE,moreau_etal2018mwe,han-etal-2020-alphamwe}. For OOV word translation for European languages, substantial improvements have been made in terms of rare and unseen words by incorporating sub-word knowledge using Byte Pair Encoding (BPE) \cite{SubwordNMT15Sennrich}. However, such methods cannot be directly applied to Chinese, Japanese and other ideographic languages. 

Integrating sub-character level information, such as Chinese ideograph and radicals  as learning knowledge has been used to enhance features in NMT systems \cite{HanKuang2018NMT,ZhangMatsumoto18radical,zhang-komachi-2018-neural}. \newcite{HanKuang2018NMT}, for example, explain that the meaning of some unseen or low frequency Chinese characters can be estimated and translated using \textit{radicals} decomposed from the Chinese characters, as long as the learning model can acquire knowledge of these radicals within the training corpus. 


Chinese characters often include two pieces of information, with \textit{semantics} encoded within radicals and a \textit{phonetic} part. The phonetic part is related to the pronunciation of the overall character, either the same or similar. 
For instance, Chinese characters with this two-stroke radical, \begin{CJK*}{UTF8}{gbsn} 刂 \end{CJK*}\,(tí dāo páng), ordinarily relate to \emph{knife} in meaning, such as the Chinese character \begin{CJK*}{UTF8}{bsmi}  劍 \end{CJK*}
 \,(jiàn, \emph{sword}) and multi-character expression \begin{CJK*}{UTF8}{bsmi} 鋒利 \end{CJK*}\,(fēnglì, \emph{sharp}). The radical \begin{CJK*}{UTF8}{gbsn} 刂 \end{CJK*}\,(tí dāo páng) preserves the meaning of knife because it is a variation of a drawing of a knife evolving from the original bronze inscription (Fig. \ref{fig:radical_knife_evolution} in Appendices).

\begin{figure*}[!t]
\begin{center}
\centering
\includegraphics*[height=1.3in]{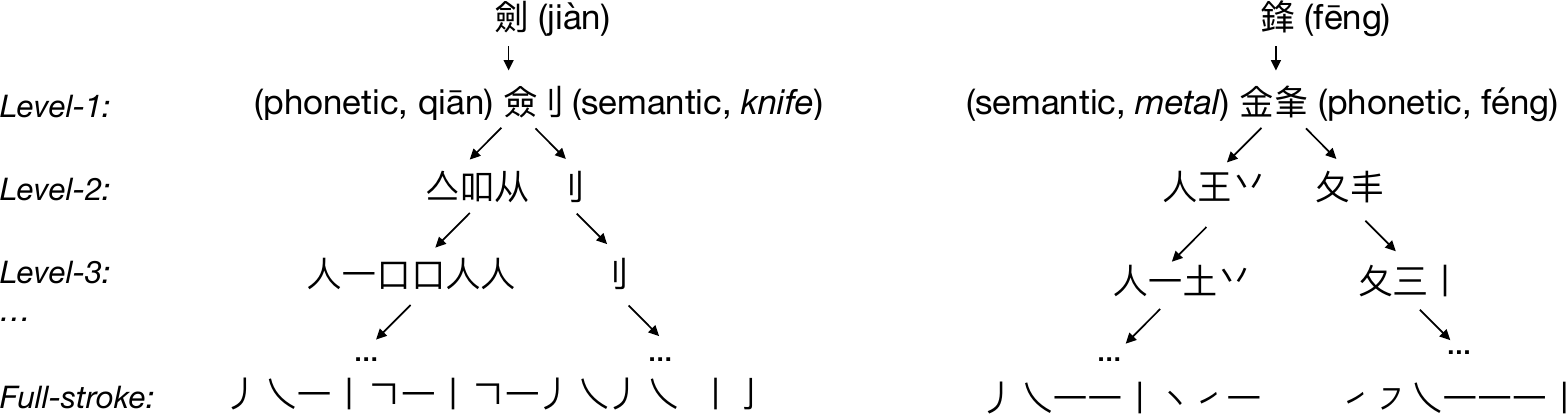}
\caption{Examples of the decomposition of Chinese characters.}
\label{fig:decompose_degree_jianfeng}
\end{center}
\end{figure*}

Not only can the radical part of a character  be decomposed into smaller fragments of strokes but the phonetic part can also be decomposed. 
Thus there are often several levels of decomposition that can be applied to Chinese characters by combining different levels of decomposition of each part of the Chinese character. As one example, Figure~\ref{fig:decompose_degree_jianfeng} shows the three decomposition levels from our model and the full stroke form of the above mentioned characters \begin{CJK*}{UTF8}{bsmi} 劍 (jiàn) and 鋒 (fēng) \end{CJK*}. 
To date, little work has been carried out to investigate the full potential of these alternative levels of decomposition of Chinese characters for the purpose of Machine Translation (MT). 

%
In this work, we  investigate Chinese character decomposition, and additionally we investigate another  area related to Chinese characters, namely Chinese MWEs.
We firstly investigate translation at increasing levels of decomposition of Chinese characters using underlying radicals, as well as the additional
Chinese character strokes (corresponding to ever-smaller units), breaking down characters into component parts as this is likely to reduce the number of unknown words. 
Then, in order to better deal with MWEs which have a common occurrence in the general context \cite{Sag2002MWE}, and working in the opposing direction in terms of meaning representation,  we investigate 
translating larger units of Chinese text, with the aim of restricting translation of larger groups of Chinese characters that should be translated together as one unit. 
In addition to investigating the effects of decomposing characters we simultaneously apply methods of incorporating MWEs into translation.  MWEs can appear in Chinese in a range of ways, such as fixed (or semi-fixed) expressions, metaphor, idiomatic phrases, and institutional, personal or location names, amongst others. 

In summary, in this paper, we investigate (i) the degree to which Chinese radical and stroke sequences represent the original word and character sequences that they are composed of; (ii) the difference in performance achieved by each decomposition level; (iii) the effect of radical and stroke representations in MWEs for MT. 
Furthermore, we offer (available at radical4mt\footnote{\url{https://github.com/poethan/MWE4MT}}): 
\begin{itemize}


\item an open-source suite of Chinese character decomposition extraction tools;

\item a Chinese $\Leftrightarrow$ English MWE corpus where Chinese characters have been decomposed 
\end{itemize}

\noindent 
The rest of this paper is organized as follows: Section 2 provides some related work in character and radical related MT;
Section 3 and 4 introduce our Chinese decomposition procedure into radical and strokes, and the experimental design; Section 5 provides our evaluations from both automatic and human perspectives; Section 6 includes  conclusions and plans for future work.

\section{Related Work}

Chinese character decomposition has been explored recently for MT.
For instance, \newcite{HanKuang2018NMT} and \newcite{ZhangMatsumoto18radical}, considered radical embeddings as additional features for Chinese $\rightarrow$ English and Japanese $\Leftrightarrow$ Chinese NMT. 
In \newcite{HanKuang2018NMT}, a range of encoding models including word+character, word+radical, and word+character+radical were tested. The final setting with word+character+radical achieved the best performance on a standard NIST \footnote{\url{https://www.nist.gov/programs-projects/machine-translation}} MT evaluation data set for Chinese $\rightarrow$ English.
Furthermore, \newcite{ZhangMatsumoto18radical} applied radical embeddings as additional features to character level LSTM-based NMT on Japanese $\rightarrow$ Chinese translation. 
None of the aforementioned work has however investigated the  performance of decomposed character sequences and the effects of varied decomposition degrees in combination with MWEs. 
Subsequently, \newcite{zhang-komachi-2018-neural} developed bidirectional English $\Leftrightarrow$ Japanese, English $\Leftrightarrow$ Chinese and Chinese $\Leftrightarrow$ Japanese NMT with word, character, ideograph (the phonetics and semantics parts of characters are separated) and stroke levels, with experiments showing that the \textit{ideograph} level was best for ZH$\rightarrow$EN MT, while the stroke level was best for JP$\rightarrow$EN MT. Although their ideograph and stroke level setting replaced the original character and word sequences, there was no investigation of 
\textit{intermediate decomposition} performance, and they only used BLEU score as the automated evaluation with no human assessment involved. 
This gives us  inspiration to explore the performance of intermediate level embedding between ideograph and strokes for the MT task.

\section{Chinese Character Decomposition}

\begin{figure}[!t]
\begin{center}
\centering
\includegraphics*[height=1.6in]{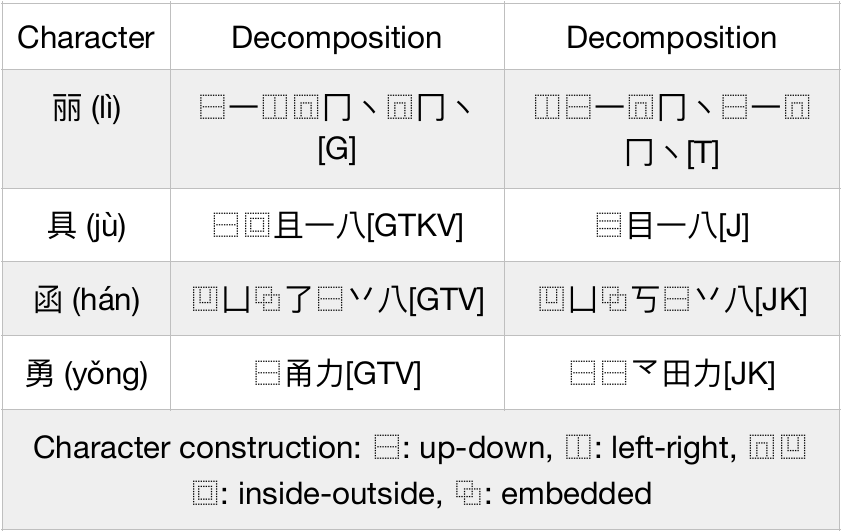}
\caption{Character examples from IDS dictionary; the grey parts of decomposition graphs represent the construction structure of the character.}
\label{fig:ids_examples}
\end{center}
\end{figure}

We introduce the character decomposition approach and the extraction tools which we apply in this work
(code will be publicly available
). We utilize the open source IDS dictionary \footnote{\url{https://github.com/cjkvi/cjkvi-ids}} 
which was derived from the CHISE (CHaracter Information Service Environment) project\footnote{\url{http://www.chise.org/}}. It is comprised of 88,940 Chinese characters from CJK (Chinese, Japanese, Korean script) Unified Ideographs and the corresponding decomposition sequences of each
character.
Most characters are decomposed as a single sequence, but characters can have up to four possible decomposed representations. The reason for this is that the character can come from different resources, such as Chinese Hanzi (G, H, T for Mainland, Hong Kong, and Taiwan), 
Japanese Kanji (J), Korean Hanja (K), and Vietnamese ChuNom (V), etc.\footnote{Universal Coded Character Set (10646:2017) \url{standards.iso.org/ittf/PubliclyAvailableStandards}} 
Even though they have the same root of Hanzi, the historical development of languages and writing systems in different territories has resulted in their certain degree of variations in the appearance and stroke order, for instances, \begin{CJK*}{UTF8}{bsmi}  (且, qiě) vs (目, mù) \end{CJK*}  from the second character example in Figure \ref{fig:ids_examples}.


Figure \ref{fig:ids_examples} shows example characters that have two different decomposition sequences. 
In our experiments, when there is more than one decomposed representation of a given character, we choose the Chinese mainland decomposition standard (G) for the model, since the corpora we use correspond best to simplified Chinese as used in mainland China. 
The examples in Figure \ref{fig:ids_examples} also show the general construction and corresponding decomposition styles of Chinese characters, such as \emph{left-right}, \emph{up-down}, \emph{inside-outside}, and \textit{embedded} 
amongst others. To obtain a decomposition level $L$ representation of Chinese character $\alpha$, we go through the IDS file $L$ times. Each time, we search the IDS file character list to match the newly generated smaller sized characters and replace them with decomposed representation recursively.

  
\section{NMT Experiments}


We test the various levels of decomposed Chinese and Chinese MWEs using publicly available data from the WMT-2018 shared tasks Chinese to English, using the preprocessed (word segmented) data as training data \cite{wmt2018findings}. The original word boundaries were preserved in decomposition sequences. To get better generalizability of our decomposition model, we used a  large  size, the first 5 million parallel sentences for training across all learning steps. 
The corpora ``newsdev2017" used for development and ``newstest2017" for testing are from WMT-2017 MT shared task \cite{wmt2017findings}. 
These include 2002 and 2001 parallel Chinese $\Leftrightarrow$ English respectively. 
We use the THUMT \cite{thumt2017} toolkit which is an implementation of several attention-based Transformer architectures \cite{google2017attention} for NMT  and set up the encoder-decoder as 7+7 layers. 
Batch size is set as 6250. For sub-word encoding BPE technology, 
we use \textit{32K BPE operations} that are learned from the bilingual training set. 
We use Google's Colab platform to run our experiments{\footnote{\url{https://colab.research.google.com}}}.
We name the baseline model using character sequences (with word boundary) as \textit{character sequence model}. For MWE integrated models, we apply the same bilingual MWE extraction pipeline from our work \cite{han-etal-2020-multimwe}, similar to \cite{rikters2017mwe},  which is an automated pre-defined PoS pattern-based  extraction procedure with filtering  threshold set to 0.85 to remove lower quality translation pairs. We integrate these extracted bilingual MWEs back into the training set to investigate if they help the MT learning.
In the decomposed models, we replace the original Chinese character sequences from the corpus with decomposed character-piece sequence inputs for training, development and testing (with original word boundary kept).

\section{Evaluation}

\begin{figure}[!t]
\begin{center}
\centering
\includegraphics*[height=0.97in]{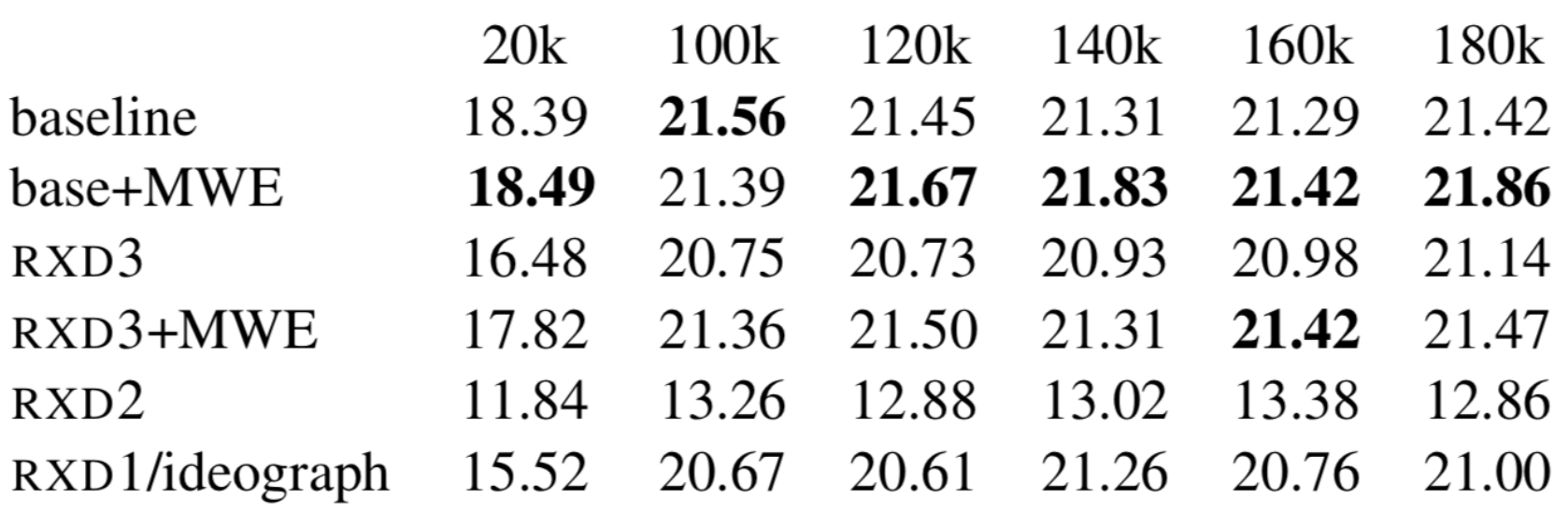}
\caption{Chinese$\rightarrow$English BLEU scores for increasing learning steps; \textsc{rxd}1/2/3 represents the decomposition level of Chinese characters. \textsc{rxd1} indicates \textit{ideograph} from \cite{zhang-komachi-2018-neural}}
\label{fig:zh2en_wmt18_bleu_xk}
\end{center}
\end{figure}

In order to assess the performance of
each model 
employing a different meaning representation in terms of decomposition and MWEs, we carried out 
both automatic, BLEU \cite{Papineni02bleu:a} in Fig. \ref{fig:zh2en_wmt18_bleu_xk}, and human evaluation (Direct Assessment) of the outputs of the system. 
Since decomposition level 3 yields generally higher scores than the other two levels, we also applied decomposition of MWEs 
to level 3 and concatenated the bilingual glossaries to the training. 

We used the models with the most learning steps, 180K, and run human evaluation on the Amazon Mechanical Turk crowd-sourcing platform,\footnote{\url{https://www.mturk.com}} including the strict quality control measures of \newcite{NLE:9961497}.
Direct Assessment scores for systems were calculated as in \newcite{translationese} by firstly computing an average score per translation before calculating the overall average for a system from its average scores for translations. Significance tests in the form of Wilcoxon Rank-Sum test are then applied to score distributions of the latter to identify systems that significantly outperform other systems in the human evaluation. 

Results of the Direct Assessment human evaluation are shown in Table~\ref{fig:sys_level_he_score} where similarly performing systems are clustered together (denoted by horizontal lines in the table).
Systems in a given lower ranked cluster are significantly outperformed by all systems in a higher ranked cluster. 
Amongst the six models included in the human evaluation, the first five form a cluster with very similar performance according to human assessors, including the baseline, MWE, \textsc{rxd}1, \textsc{rxd}3MWE, and \textsc{rxd}3 which do not outperform each other with any significance.
\textsc{rxd2}, on the other hand, is far behind the other models in terms of performance according to human judges (also the automated BLEU score) performing significantly worse than all other runs (at $p<$  0.05). As the tradition of WMT shared task workshop, we cluster the first five models into one group, while the \textsc{rxd}2 into a second group.
Furthermore, human evaluation results in Table~\ref{fig:sys_level_he_score} show that the top five models all achieve high performance on-par with state-of-the-art in Chinese to English MT.

We also discovered that the decomposed models generated fewer system parameters for the neural nets to learn, which potentially reduces  computational complexity. 
For instance, the total trainable variable size of the character sequence baseline model is 89,456,896, while this number decreased to 80,288,000 and 80,591,104 respectively for the \textsc{rxd}3 and \textsc{rxd}2 models (a 10.25\% drop for \textsc{rxd}3). As mentioned by \newcite{Goodfellow-et-al-2016}, in NLP tasks the total number of possible words is so large that the word sequence models have to operate on an extremely high-dimensional and sparse discrete space. 
The decomposition model reduced the overall size of possible tokens for the model to learn, which is more space efficient.


\begin{table}[!t]
\begin{center}
\centering
\begin{tabular}{crccc}
\toprule
\multicolumn{1}{c}{Ave.} 
     & \multicolumn{1}{c}{Ave. $z$}     
                & $n$   & $N$   & \\
\multicolumn{1}{c}{raw} 
     & \multicolumn{1}{c}{}     
                &    &   & \\
\midrule
73.2 & 0.161    & 1,232 & 1,639 & \textsc{base}    \\
71.6 & 0.125    & 1,262 & 1,659 & \textsc{MWE}     \\
71.6 & 0.113    & 1,257 & 1,672 & \textsc{rxd1}    \\ 
71.3 & 0.109    & 1,214 & 1,593 & \textsc{rxd3MWE} \\ 
70.2 & 0.073    & 1,260 & 1,626 & \textsc{rxd3}    \\ \hline
53.9 & $-$0.533 & 1,227 & 1,620 & \textsc{rxd2}    \\  
\bottomrule
\end{tabular}
\caption{Human evaluation results for systems using Direct Assessment, where Ave. raw $=$ the average score for translations calculated from raw Direct Assessment scores for translations, Ave. $z$ $=$ the average score for translations after score standardization per human assessor mean and standard deviation score, $n$ is the number of distinct translations included in the human evaluation (the sample size used in significance testing), $N$ is the number of human assessments (including repeat assessment).}
\label{fig:sys_level_he_score}
\end{center}
\end{table}



For the automatic and human evaluation results, where the decomposition level 2 achieved surprisingly lower score than the other levels, error analysis revealed an important insight. While level-1 decomposition encoded the original character sequences into radical representations, and this typically contains semantic and phonetic parts of the character, level-3 gives a deeper decomposition of the character such as the stroke level pieces with sequence order. In contrast, however, level-2 decomposition appears to introduce some intermediate characters that mislead model learning. These intermediate level characters are usually constructed from fewer strokes than the original root character, but can be decomposed from it. As in Figure \ref{fig:decompose_degree_jianfeng}, from decomposition level-2, we get new characters \begin{CJK*}{UTF8}{gbsn} 从\,(cóng) and 王\,(wáng) respectively from  \end{CJK*} \begin{CJK*}{UTF8}{bsmi} 劍\,(Jiàn, \textit{sword}) and 鋒 \end{CJK*}\,(fēng, \textit{edge/sharp point}), but they have no direct meaning from their father characters, instead meaning ``from" and ``king" respectively. 
In summary, decomposition level-2 tends to generate some intermediate characters that do not preserve the meaning of the original root character's radical, nor those of the strokes, but rather smaller sized independent characters with fewer strokes that result in other meanings.


\section{Conclusions and Future Work}
In this work, we tested the varying degrees of Chinese character decomposition and their effect on Chinese to English NMT with attention architecture. To the best of our knowledge, this is the first work on detailed decomposition level of Chinese characters for NMT, and decomposition representation for MWEs. 
We conducted experiments for decomposition levels 1 to 3; we had a look at level 4 decomposition and it appears similar to level 3 sequences. We publish our extraction toolkit free for academic usage.  We conducted automated evaluation with the BLEU metric, and crowd sourced human evaluation with the direct assessment (DA) methodology. 
Our conclusion is that the Chinese character decomposition levels 1 and 3 can be used to represent or replace the original character sequence in an MT task, and that this achieves similar performance to the original character sequence model in our NMT setting. However, decomposition level 2 is not suitable to represent the original character sequence in meaning at least for MT. We leave it to future work to explore the performance of different decomposition levels in other NLP tasks.

Another finding from our experiments is that while adding bilingual MWE terms can both increase character and decomposed level MT score according to the automatic metric BLEU, the human evaluation shows no statistical significance between them. 
Significance testing using automated evaluation metrics will be carried out in our future work, such as METEOR \cite{BanerjeeLavie2005meteor}, and LEPOR \cite{han2012lepor,han2014lepor}, in addition to BLEU.

We will consider different MWE integration methods in future and reduce the training set to investigate the differences in low-resource scenarios (5 million sentence pairs for training set were used in this work). We will also sample a set of the testing results and conduct a human analysis regarding the MWE translation accuracy from different representation models.
We will further investigate different strategies of \textit{combining} several level of decompositions together and their corresponding performances in semantic representation, such as MT task. 
The IDS file we applied to this work limited the performance of full stroke level capability, and we will look for alternative methods to achieve full-stroke level character sequence extraction for NLP tasks investigation.

\section*{Acknowledgments}
We thank Yvette Graham for helping with human evaluation, Eoin Brophy for helps with Colab, and thank the anonymous reviewers for their thorough reviews and insightful feedback.
The ADAPT Centre for Digital Content Technology is funded under the SFI Research Centres Programme (Grant 13/RC/2106) and is co-funded under the European Regional Development Fund. The input of Alan Smeaton is part-funded by Science Foundation Ireland 
under grant number SFI/12/RC/2289 (Insight Centre). 



\bibliographystyle{acl_natbib}
\bibliography{nodalida2021}

\section*{Appendices}


\subsection*{Appendix A: Chinese Character Knowledge}
\begin{figure*}[!t]
\begin{center}
\centering
\includegraphics*[height=1.5in]{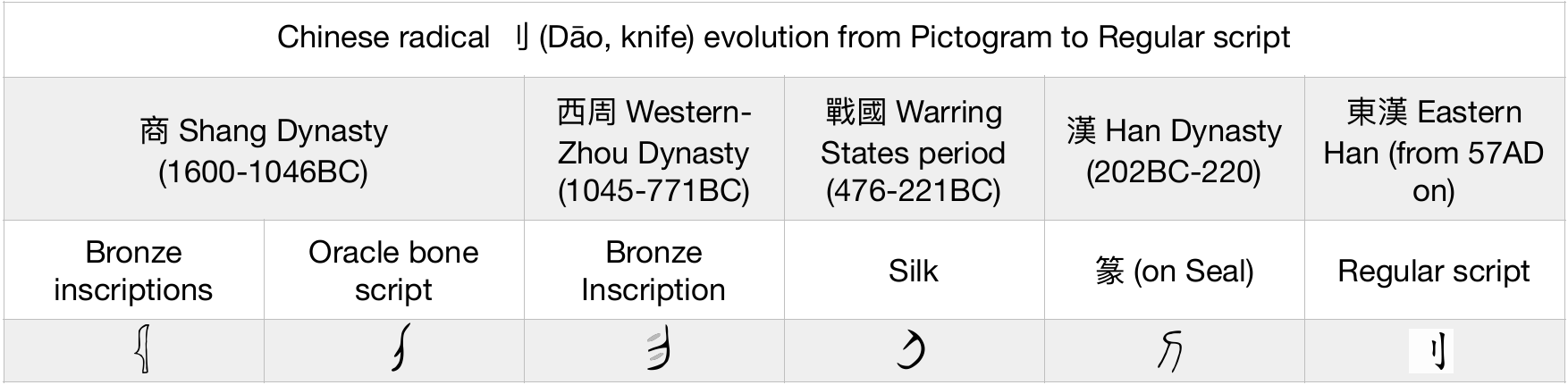}
\caption{Example Chinese radical, \begin{CJK*}{UTF8}{gbsn} 刂 \end{CJK*} (Dāo), where the character evolved from leftmost pictogram to present day regular script (rightmost) containing only two strokes. The two strokes are called as \begin{CJK*}{UTF8}{bsmi} 豎\, (Shù, vertical) + 豎鈎 \end{CJK*}\, (Shù gōu, vertical with hook). The corresponding character representation is \begin{CJK*}{UTF8}{gbsn} 刀\end{CJK*}\,(Dāo).} 
\label{fig:radical_knife_evolution}
\end{center}
\end{figure*}

Figure \ref{fig:radical_knife_evolution} demonstrates the meaning preservation root of Chinese radicals,  where 
the evolution of the Chinese character \begin{CJK*}{UTF8}{gbsn} 刀\,(Dāo), meaning \emph{knife}, evolved from bronze inscription form to contemporary character and radical form, 刂\,(named as: tí dāo páng).\end{CJK*} 

\begin{figure*}[!t]
\begin{center}
\centering
\includegraphics*[height=1.0in]{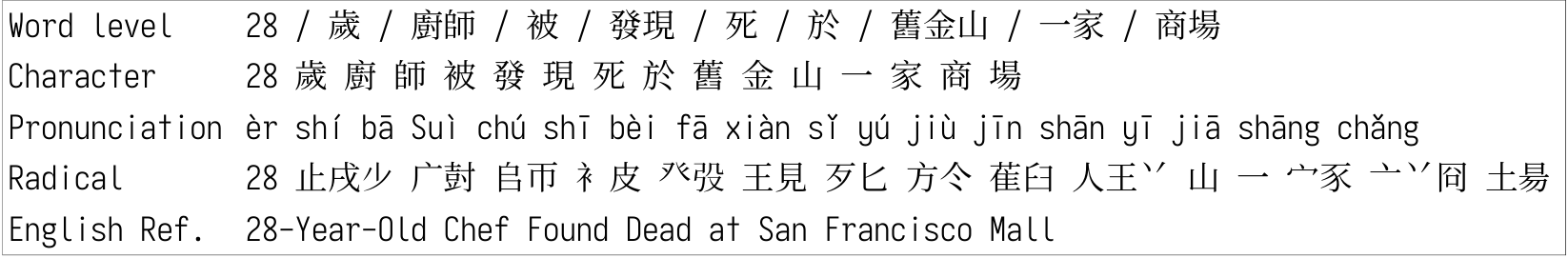}
\caption{Example of Chinese word to character level changes for MT. Pronunciation is Mandarin in Pinyin. The English reference here is taken from
the corpus we used for our experiments.}
\label{fig:word2character_MT}
\end{center}
\end{figure*}

NMT for Asian languages has included translation at the level of phrase, word, and character sequences (see Figure \ref{fig:word2character_MT}). 

\subsection*{Appendix B: More Details of Evaluation}

The evaluation scores of character sequence baseline NMT, character decomposed NMT and MWE-NMT according to the BLEU metric are presented in Fig. \ref{fig:zh2en_wmt18_bleu_xk}. 
The \textsc{rxd1} model, decomposition level 1, is the \textit{ideograph} model \newcite{zhang-komachi-2018-neural} used for their experiments where the phonetics \begin{CJK*}{UTF8}{gbsn} (声旁\,shēng páng) and semantics (形旁\, xíng páng)  \end{CJK*}parts of character are separated initially.

\begin{figure*}
\begin{center}


{\centering
\begin{tikzpicture}
\begin{axis}
[%
graph,
height=\graphheight,
width=\graphwidth,
xlabel=Learning Steps,
ylabel=BLEU scores,
ymin=11,
legend style={
cells={anchor=west},
style={at={(1.0,0.8)}}
}
]
\pgfplotstableread[col sep=comma]{zh2en_NMT_BLEU_scores.txt}\plotdata

\addplot[color=aqua,mark=triangle] table[x=models, y=baseline] {\plotdata};
\addlegendentry{\textsc{base}}

\addplot[color=blue,mark=*] table[x=models, y=rxd1] {\plotdata};
\addlegendentry{\textsc{rxd1}}

\addplot[color=black,mark=o] table[x=models, y=rxd2] {\plotdata};
\addlegendentry{\textsc{rxd2}}

\addplot[color=red,mark=square] table[x=models, y=rxd3] {\plotdata};
\addlegendentry{\textsc{rxd3}}

\addplot[color=yellow,mark=diamond] table[x=models, y=rxd3+MWE] {\plotdata};
\addlegendentry{\textsc{rxd3}MWE}

\end{axis}
\end{tikzpicture}

}
\caption{Learning curves from different models with BLEU metric}\label{steps}
\end{center}
\end{figure*}
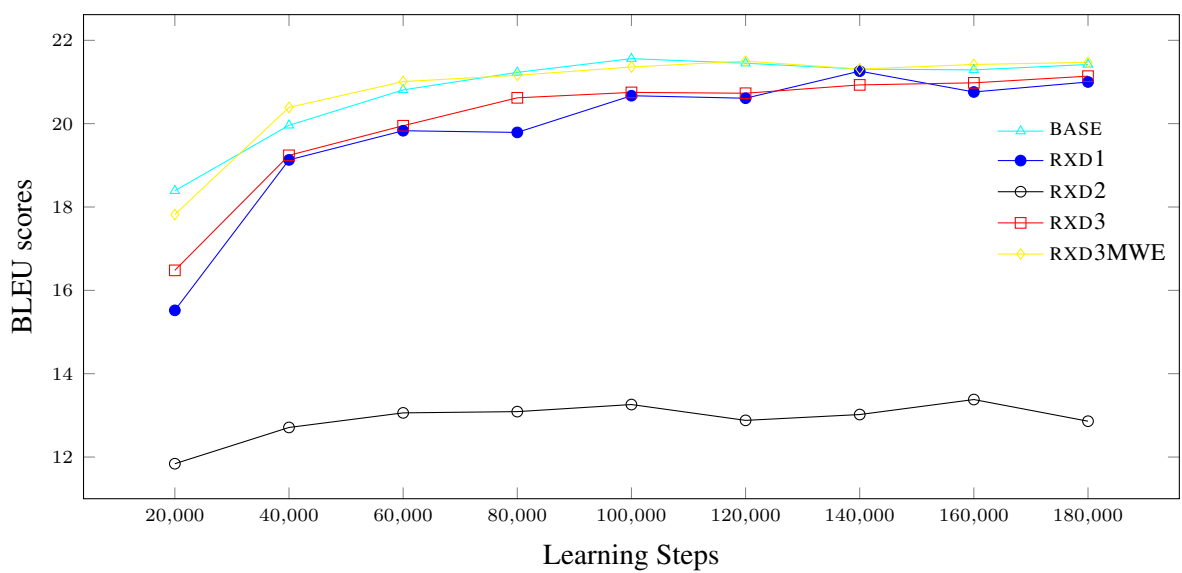

From the automated evaluation results, we see that decomposition model \textsc{rxd}3 has very close BLEU scores to the baseline character sequence (both with word boundary) model. This is very interesting since the level 3 Chinese decomposition 
is typically impossible (or too difficult) for even native language human speakers to read and understand. 
Furthermore, by adding the decomposed MWEs back into the learning corpus, ``rxd3+MWE" (\textsc{rxd3}MWE) yields higher BLEU scores in some  learning steps than the baseline  model. 
To gain further
insight, we provide the learning curve  with the learning steps and corresponding automated-scores in Figure \ref{steps}.

The BLEU score increasing ratio in decomposed models (from \textsc{rxd}3 to \textsc{rxd}3MWE) is larger than the ratio in original character sequence models (from \textsc{base} to \textsc{base}MWE) by adding MWEs in general. 
Furthermore, the increase in performance is very consistent by adding MWEs from the decomposed model, compared to the conventional character sequence model. For instance, the performance has a surprisingly drop at 100K learning steps for \textsc{base}MWE. 

\subsection*{Appendix C: Looking into MT Examples}

From the learning curves in Fig. \ref{steps}, we suggest  that with 5 million training sentences and 7+7 layers of encoder-decoder neural nets, the Transformer model becomes too flat in its learning rate curve with 100K learning steps, and this applies to both original character sequence model and decomposition models.

\begin{figure*}[!t]
\begin{center}
\centering
\includegraphics*[height=3.4in]{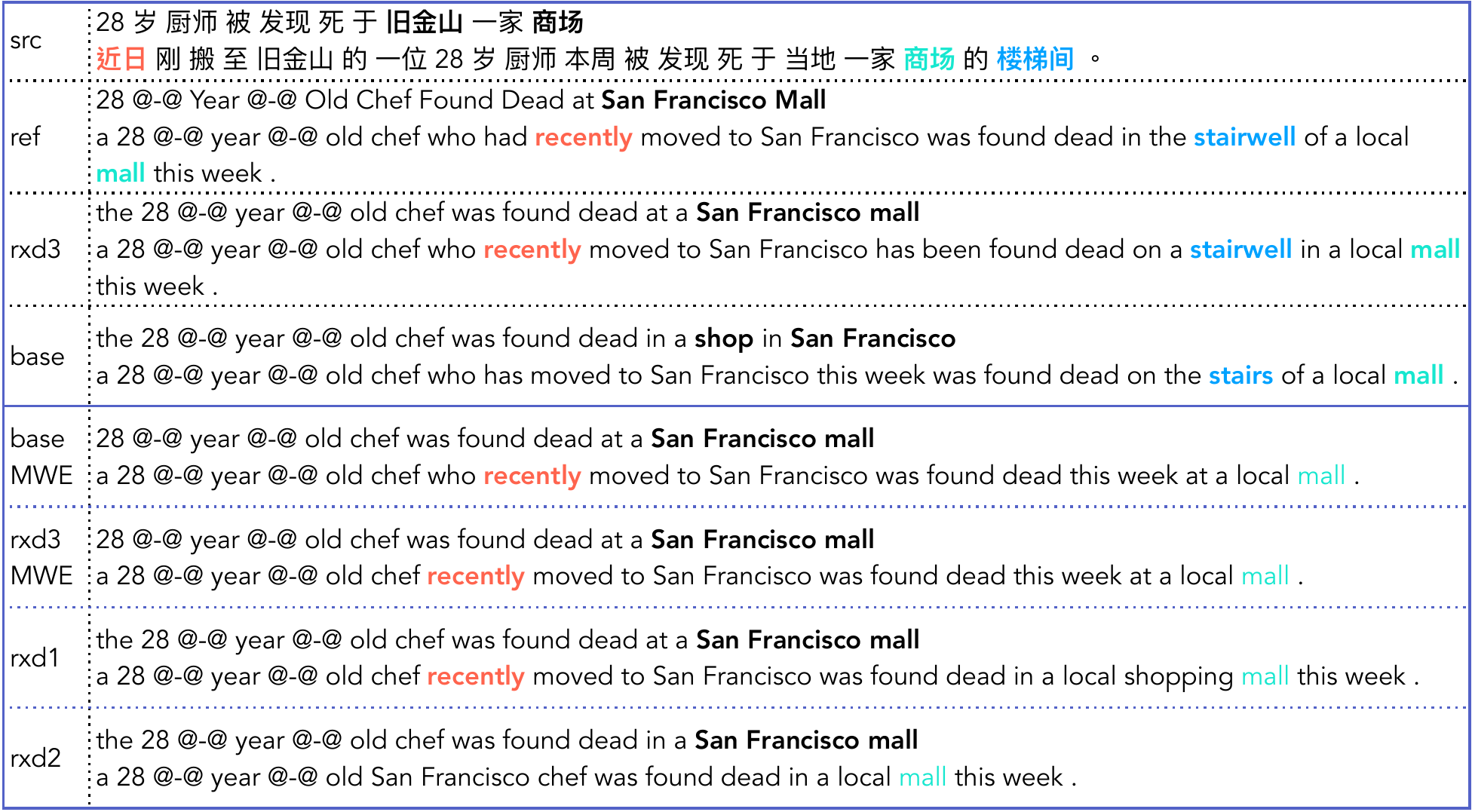}
\caption{Samples of the English MT output at 100K learning steps: \textsc{rxd1}, \textsc{rxd2} and \textsc{rxd3} are the Chinese decomposition with level 1 to 3, \textsc{base} is the character sequence model, \textsc{base}MWE and \textsc{rxd3}MWE are character sequence model with MWEs and decomposition level 3 model with decomposed MWEs, and src/ref represents source/reference.}
\label{fig:MToutput_rxd3_vs_base_100k}
\end{center}
\end{figure*}

In light of this, we look at the MT outputs from head sentences of testing file at 100K learning steps models, and provide some insight into errors made by each model. Even though the automated BLEU metric gives the baseline model a higher score 21.56 than the \textsc{rxd3} model (20.75) the translation of some Chinese MWE terms is better with the \textsc{rxd3} model. 
For instance, in Figure \ref{fig:MToutput_rxd3_vs_base_100k}, the Chinese MWE \begin{CJK*}{UTF8}{gbsn} 商场\,(Shāngchǎng) in the first sentence is correctly translated as \emph{mall} by \textsc{rxd3} model but translated as \emph{shop} by the baseline character sequence model; the MWE 楼梯间\,(lóutījiān) in the second sentence is correctly translated as \emph{stairwell} by the \textsc{rxd3} model while translated as \emph{stairs} by baseline. Furthermore, the MWE 近日\,(Jìnrì) meaning \textit{recently} is totally missed out by the original character sequence model, which results in a misleading ambiguous translation of an even larger content, i.e., did the chief moved to San Francisco (SF) \textit{recently} or \textit{this week}. We will not get this clearly from the character base sequence model, however, the MWE 近日 \end{CJK*}\,(Jìnrì) is correctly translated by the  \textsc{rxd3} model and the overall meaning of the sentence is clear that the chef moved to SF \textit{recently} and was found dead \textit{this week}. 

We also attach the translations of these two sentences by  four other models. With regard to the first sentence MWEs, all the four models translate San Francisco mall correctly as \textsc{ref} and \textsc{rxd3} beating \textsc{base} model. In terms of the second sentence MWEs, \textsc{base}MWE and \textsc{rxd}2 drop out the MWE \begin{CJK*}{UTF8}{gbsn} 近日\,(Jìnrì, \textit{recently}) as \textsc{base} model, and all the four models drop out the translation of MWE 楼梯间 \end{CJK*}\,(lóutījiān, \textit{stairwell}).




\end{document}